\documentclass{article}

\usepackage{microtype}
\usepackage{graphicx}
\usepackage{subfigure}
\usepackage{enumitem}
\usepackage{booktabs} 

\usepackage{hyperref}

\usepackage[accepted]{icml2024}

\usepackage{amsmath}
\usepackage{amssymb}
\usepackage{mathtools}
\usepackage{amsthm}

\usepackage{pgfplots}

\usepackage[capitalize,noabbrev]{cleveref}

\theoremstyle{plain}

\theoremstyle{definition}

\theoremstyle{remark}

\usepackage[textsize=tiny]{todonotes}
\usepackage{multirow}

\icmltitlerunning{Policy Induction: Predicting Startup Success via Explainable Memory-Augmented In-Context Learning}

\begin{document}

\twocolumn[
\icmltitle{Policy Induction: Predicting Startup Success via Explainable Memory-Augmented In-Context Learning}



\icmlsetsymbol{equal}{*}

\begin{icmlauthorlist}
\icmlauthor{Xianling Mu}{yyy}
\icmlauthor{Joseph Ternasky}{comp}
\icmlauthor{Fuat Alican}{comp}
\icmlauthor{Yigit Ihlamur}{comp}

\begin{center}
\footnotesize
\textsuperscript{(1)} University of Oxford \quad
\textsuperscript{(2)} Vela Research
\end{center}
\vskip 0.15in

\end{icmlauthorlist}

\icmlaffiliation{yyy}{University of Oxford, Oxford, United Kingdom}
\icmlaffiliation{comp}{Vela Research, San Francisco, United States}
\icmlaffiliation{comp}{Vela Research, San Francisco, United States}
\icmlaffiliation{comp}{Vela Research, San Francisco, United States}

\icmlcorrespondingauthor{Xianling Mu}{xianling.mu@chch.ox.ac.uk}
\icmlcorrespondingauthor{Yigit Ihlamur}{yigit@vela.partners}

\icmlkeywords{Machine Learning, ICML}

\vskip 0.3in
]


\begin{abstract}
Early-stage startup investment is a high-risk endeavor characterized by scarce data and uncertain outcomes. Traditional machine learning approaches often require large, labeled datasets and extensive fine-tuning, yet remain opaque and difficult for domain experts to interpret or improve. In this paper, we propose a transparent and data-efficient investment decision framework powered by memory-augmented large language models (LLMs) using in-context learning (ICL). Central to our method is a natural language policy embedded directly into the LLM prompt, enabling the model to apply explicit reasoning patterns and allowing human experts to easily interpret, audit, and iteratively refine the logic. We introduce a lightweight training process that combines few-shot learning with an in-context learning loop, enabling the LLM to update its decision policy iteratively based on structured feedback. With only minimal supervision and no gradient-based optimization, our system predicts startup success far more accurately than existing benchmarks. It is over 20$\times$ more precise than random chance, which succeeds 1.9\% of the time. It is also 7.1$\times$ more precise than the typical 5.6\% success rate of top-tier venture capital (VC) firms.
\end{abstract}

\section{Introduction}
ICL has emerged as a powerful paradigm in natural language processing, offering several key advantages: unsupervised adaptability, minimal data requirements, and strong reasoning capabilities \citep{dong2023survey}. These features make ICL well suited for domains such as early-stage VC, where high-stakes decisions must often be made with sparse, noisy, or incomplete data.

In this paper, we introduce a novel investment decision-making framework that integrates memory-augmented LLMs with ICL. At the core of our system is a natural-language ``policy''---a structured set of heuristics expressed in plain text---which is embedded directly into the LLM prompt. These policies guide the model’s reasoning during prediction while remaining fully interpretable and editable by human experts.

We begin by prompting the LLM with a small set of labeled examples (successful and failed startups) to generate an initial decision policy. This policy is then iteratively refined through a lightweight training loop: new examples are incorporated as additional context, prompting the model to revise and improve the policy. At each step, candidate policies are scored using a precision-based evaluation metric, and the highest-performing version is retained. To further enhance policy quality, we incorporate both automated reflection---LLM-generated explanations of prediction outcomes---and optional expert intervention. This iterative refinement process continues until a stable, high-performing policy emerges that generalizes well to unseen data.

Our primary contributions are as follows:
\begin{itemize}
    \item \textbf{Transparency and Interpretability:} All model logic is encoded in plain-text policies that are human-readable and editable. This allows VC experts to understand, audit, and improve the model’s decision-making process — a critical requirement in high-stakes financial contexts.

    \item \textbf{Efficiency and Cost-Effectiveness:} Our method requires minimal fine-tuning, short training cycles, and very low compute cost. Using only the GPT-4o mini API and a few dollars' worth of compute, we are able to produce policies that outperform random baselines by 3–4$\times$ in precision, even without any further optimization.

    \item \textbf{Generalizability and Transferability:} Because our decision policies are expressed in natural language, they can be seamlessly applied across tasks without requiring model retraining or code changes. This makes the approach highly transferable to other domains where structured reasoning is essential, such as grant evaluation, academic hiring, or legal case review. The policy format allows domain experts to adapt, inspect, and reuse decision logic with minimal effort, enabling broad applicability beyond the startup investment context.
\end{itemize}

\noindent The remainder of this paper details our related work, methodology, dataset, experiments, and findings, demonstrating that our approach yields practical, accurate, and interpretable investment predictions.

\section{Related Work}
\paragraph{ICL.}
ICL has emerged as a core capability of LLMs, allowing models to generalize to new tasks by conditioning on examples without requiring parameter updates. Recent studies have shown that ICL can support algorithmic reasoning \citep{zhou2022least, wang2023selfconsistency} and instruction following \citep{wei2021finetuned}. Our work builds on this growing body of research by exploring how ICL can support iterative refinement through natural language policies. Rather than introducing new model architectures, we aim to make this process interpretable and accessible by using plain-text reasoning strategies that evolve over time.

\paragraph{LLMs for Decision Support.}
LLMs are increasingly being applied to structured decision-making tasks such as fairness-aware hiring evaluations. Many of these applications leverage the language modeling strength of LLMs but may rely on static prompts or produce decisions that are difficult to audit. In this context, we contribute a simple framework for integrating explicit decision heuristics in natural language, which may help improve transparency and controllability in domains such as early-stage venture evaluation.

\paragraph{Predicting Startup Success.}
Previous research on startup success prediction has used structured financial, operational, and social network data, often via traditional machine learning classifiers. While these methods offer useful baselines, they typically depend on predefined features and fixed decision logic. More recent efforts have explored the adaptation of LLMs to the VC domain \citep{xiong2024GPTree}. Our approach attempts to complement these methods by exploring how LLMs can reason from structured founder profiles using explicitly defined policies that are easy to inspect and modify.

\paragraph{Memory-Augmented Models.}
Our approach is loosely inspired by memory-augmented neural architectures \citep{graves2016hybrid}, which enhance reasoning by maintaining dynamic external state. Rather than altering internal representations, we simulate a form of symbolic memory by updating natural language policies across training iterations. This design allows LLMs to refine their behavior using plain-text cues alone, an approach that emphasizes interpretability over architectural complexity.

\section{Dataset}
\subsection{Dataset Overview}
All data used in this study originate from a dataset we refer to as \texttt{founder\_cleaned\_data}, which was constructed by converting unstructured information from LinkedIn profiles and Crunchbase entries into structured features using LLM-powered extraction techniques.

This dataset focuses on US-based companies founded in or after 2010 and contains information on 1,022 successful and 9,902 failed companies. A company is labeled as \textit{successful} if it has completed an initial public offering (IPO) with a valuation over \$500M, has been acquired for more than \$500M, or has raised more than \$500M and is still operating. A company is labeled as \textit{failed} if it was founded between 2010 and 2020 and raised between \$100K and \$4M. It must still be operating, but show no reasonable signs of a successful outcome.

Each entry includes the primary structured features used for analysis, notably: founder characteristics such as \texttt{clean\_cb\_profile}, \texttt{clean\_linkedin\_profile}, \texttt{company\_description}, and \texttt{idea}; and outcome labels such as \texttt{funds\_range} and \texttt{success}.

To mitigate data contamination in LLMs, as discussed in \citet{palavalli2024leakage}, we excluded fields that contain potentially identifiable information, such as company and founder names. Therefore, we used only \texttt{clean\_cb\_profile}, \texttt{clean\_linkedin\_profile} and \texttt{success} for model training and evaluation, ensuring that the model could not ``cheat'' by memorizing known success cases.

\subsection{Training Data}
We initially allocated 200 successful and 200 unsuccessful founders as a candidate training set. However, in most experiments, we did not use the full set. The best-performing policy was trained using only 120 successful and 120 unsuccessful examples.

In practice, we observed that as few as 60 successful and 60 unsuccessful examples were sufficient to generate reasonably strong policies. This suggests that our method is data-efficient and benefits significantly from the ICL capabilities of LLMs.

\subsection{Testing Data}
\label{sec:testing-data}
All remaining data were reserved for validation and testing. We employed two types of test set configurations to evaluate policy performance. The first consisted of 100 successful and 1000 unsuccessful cases, used for preliminary assessment and comparison across different policies. The second configuration included 40 successful and 2000 unsuccessful cases, designed to simulate a more realistic success rate comparable to the observed proportion of successful founders.

\section{Methodology}
Our core methodology focuses on generating and optimizing a policy to assist an LLM in predicting the success of startup founders. The prediction task itself is entirely delegated to the LLM, with the policy embedded in the prompt to guide its reasoning.

A typical inference prompt follows this structure:

\begin{quote}
\ttfamily
You are an expert in venture capital, specializing in evaluating startup founders.
Your task is to distinguish successful founders from unsuccessful ones. \\
Here is a policy to assist you: \{policy\}\\

Given the founder's profile:\\
Founder's LinkedIn Profile: \{row['clean\_linkedin\_profile']\}\\
Crunchbase information: \{row['clean\_cb\_profile']\}\\

Based on this information, determine if the founder will succeed. Answer using only one word: True or False.
\end{quote}

The complete training pipeline, as illustrated in the figure, comprises three major components:

\begin{itemize}
    \item {Initial Policy Generation}
    \item {In-Context Learning Loop for Iterative Policy Refinement}
    \item {Further Policy Enhancement via Reflection or Expert Intervention}
\end{itemize}

One of the key benefits of our framework is its flexibility: policy improvement does not rely on a strict sequential pipeline. Instead, various methods can be applied iteratively or in combination, with each resulting policy evaluated and \hyperref[sec:scoring]{scored}. The highest-scoring policy is selected as the final output.

\vspace{-1em}
\begin{figure}[h]
    \centering
    \includegraphics[width=0.48\textwidth]{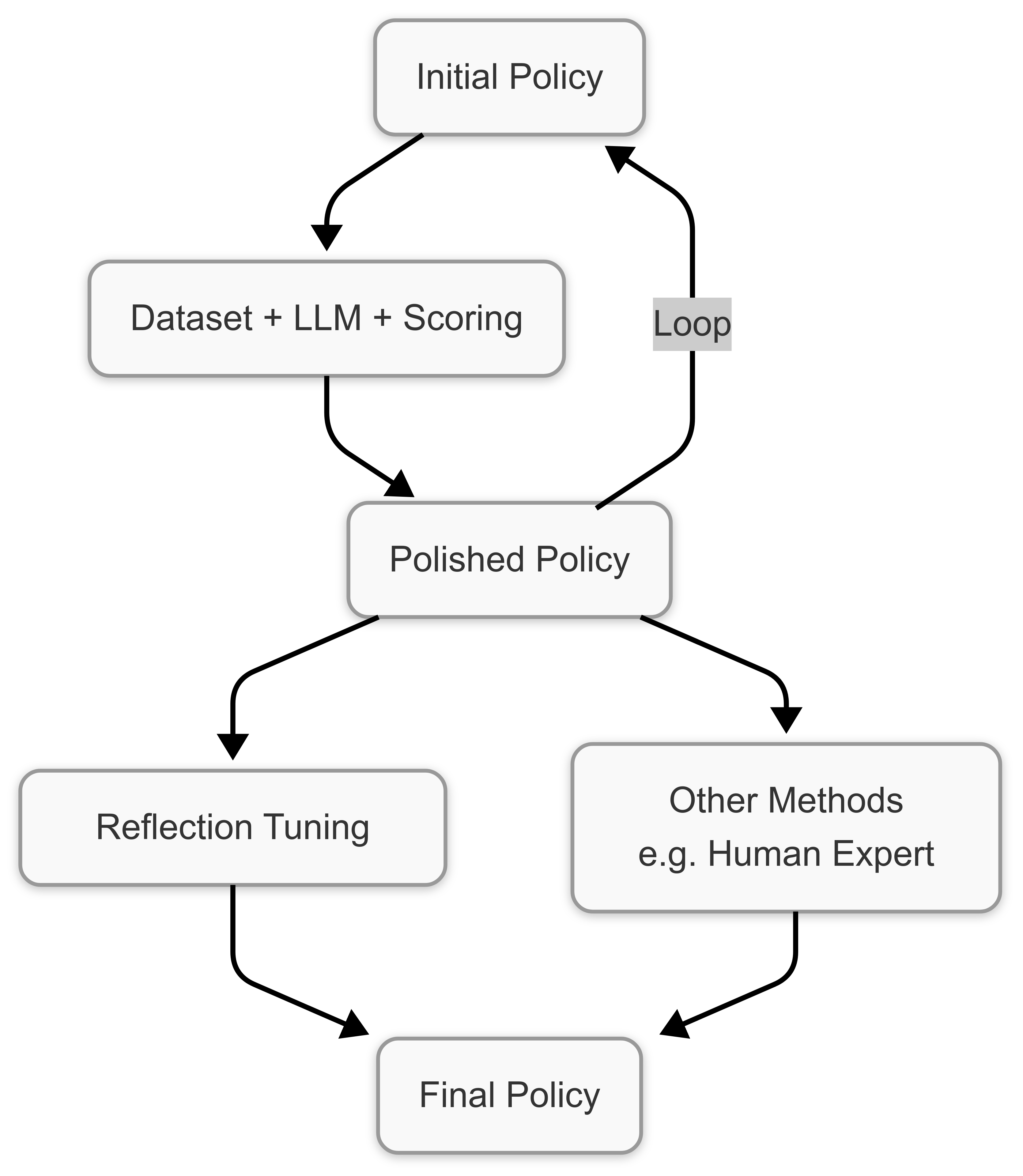}
    \caption{Policy Generating Workflow}
    \label{fig:example}
\end{figure}

\subsection{Initial Policy Generation}

We generated the initial policy using a prompt-based approach, leveraging 20 successful and 20 unsuccessful cases combined with expert-informed editing. The LLM was tasked with producing a structured set of rules based on this minimal dataset. An example of such a refined initial policy is as follows:

\textit {Refined Policy for Distinguishing Successful Founders:}
\begin{enumerate}[itemsep=0.3em, topsep=0em]
    \item \textit{Educational Background — Advanced degrees (MD, PhD, MBA) from reputable institutions are preferred, showcasing expertise in relevant fields.}
    \item \textit{Industry Experience — Extensive experience in relevant industries (biotechnology, healthcare) with leadership roles in both academic and operational capacities.}
    \item \textit{Professional Network — A robust LinkedIn network (500+ connections), indicating strong engagement and relationships within the biotech community.}
    \item \textit{Founding Experience — Founders should have previous entrepreneurial initiatives or significant contributions to new ventures.}
    \item \textit{Technical and Scientific Competence — Strong technical skills and a proven record of innovation in relevant scientific fields (e.g., drug discovery, mRNA technology).}
    \item \textit{Advisory and Leadership Roles — Recognized as a leader with experience in advisory or governance capacities in prominent organizations.}
    \item \textit{Visibility in the Industry — Engagement in key projects that enhance credibility and professional visibility.}
    \item \textit{Collaborative and Interdisciplinary Work — Demonstrated ability to foster collaborations between academia and industry or across various sectors.}
    \item \textit{Investment and Funding Experience — Proven track record in securing funding or navigating financial structuring for biotech ventures.}
\end{enumerate}

\subsection{In-Context Learning Loop}
\label{sec:icl-loop}

This forms the core of our training strategy. For each training data point, we prompt the LLM to incorporate new information into the existing policy. A sample prompt is as follows:

\begin{quote}
\ttfamily
You are an expert in venture capital tasked with distinguishing successful founders from unsuccessful ones.
Based on past experience, you have established the following policy: \{policy\}.\\

Recently, a new case was discovered:
Founder's LinkedIn Profile: \{row['clean\_linkedin\_profile']\}\\
Crunchbase information: \{row['clean\_cb\_profile']\}\\
The founder was eventually unsuccessful.\\

Summarize this new case to refine and expand the existing policy.
Provide only the updated policies as your response.
Your policy should be well-structured and have fewer than 20 rows.
\end{quote}

We implemented two complementary strategies:

\begin{itemize}
    \item \textbf{Sequential Update:} Iterate through each training data point. At each step, generate a new policy from the current policy and the new data. Evaluate both the original and the new policy using a scoring system, and always keep the one with the higher score.

    \item \textbf{Parallel Update with Selection:} For each training data point, simultaneously generate a new policy from the current policy and the data. All policies are scored, and the top 10\% are selected as effective examples. Then a new policy is generated on the basis of these high-quality samples.
\end{itemize}

Each strategy has its strengths. The parallel method is faster and performs better in the early stages of training. However, the sequential method typically produces better final policies when aiming for maximum performance.

These two strategies can also be used in a looped fashion to form an iterative refinement cycle. For example, after generating new candidate policies from training examples and scoring them, we can:

\begin{enumerate}[itemsep=0.2em, topsep=0.2em]
    \item Select the best-performing policy,
    \item Feed it back into another round of sequential or parallel updates using fresh or prioritized examples,
    \item Re-score and repeat the process.
\end{enumerate}

This looped architecture enables continual improvement of the policy through multiple passes over the data, allowing the model to integrate increasingly refined distinctions and patterns from both successful and failed cases.

\subsection{Further Refinement}

After applying the ICL loop, the resulting policies are already strong. However, we further enhance performance by integrating reflections and expert edits.

For selected representative examples, we prompt the LLM to generate a one-sentence reflection on why a given founder succeeded or failed:

\begin{quote}
\ttfamily
You are an expert in venture capital, specializing in evaluating startup founders.\\
Your task is to distinguish successful founders from unsuccessful ones.\\

Given the founder's profile:\\
Founder's LinkedIn Profile: \{row['clean\_linkedin\_profile']\}\\
Crunchbase information: \{row['clean\_cb\_profile']\}\\
The founder was eventually \{success\_status\}.\\

In ONE SENTENCE, using the founder's background,\\
clearly explain the key reason why this founder \{success\_verb\}.
\end{quote}

These reflections are then used to revise and expand the policy. In addition, human experts can manually modify or reorder the policy rules based on domain expertise, providing an interpretable, high-performance policy ready for deployment.

\section{Experiments}

To evaluate the effectiveness of our approach, we conducted experiments using a fixed test set composed of 1,000 failed founders and 100 successful founders as the standard test set. This reflects a natural precision baseline of 9.09\%, which aligns with the natural precision observed in our dataset and facilitates the acquisition of additional test data. Given the nature of the investment domain, where capital efficiency and outcome quality are paramount, we prioritize precision over recall and use the $F_{0.5}$ score as our primary evaluation metric.

For all evaluations, the LLM was prompted with the subject's \texttt{clean\_cb\_profile} and \texttt{clean\_linkedin\_profile}, along with the decision-making policy where applicable. The model was tasked with making a binary prediction (\texttt{True} or \texttt{False}) for each founder's likelihood of success.

\subsection{Vanilla Test}

In the vanilla test, we prompted several LLMs without any policy guidance, relying solely on the internal reasoning capabilities of the model. Importantly, this setup was identical to our complete prompting pipeline except that the decision policy was removed. All other components, including input fields, output format, and overall structure, remained unchanged. The prompt structure was as follows:

\begin{quote}
\ttfamily
You are an expert in venture capital, specializing in evaluating startup founders.\\
Your task is to distinguish successful founders from unsuccessful ones.\\

Given the founder's profile:\\
Founder's LinkedIn Profile: \{row['clean\_linkedin\_profile']\}\\
Crunchbase information: \{row['clean\_cb\_profile']\}\\

Based on this information, determine if the founder will succeed.\\
Answer using only one word: True or False
\end{quote}

The performance of three different LLMs: GPT-4o-mini, GPT-4o and the most powerful o3 is summarized below:

\begin{table}[h!]
\centering
\begin{tabular}{l c c c c}
\toprule
Model & Accuracy & Precision & Recall & $F_{0.5}$ \\
\midrule
GPT-4o-mini & 0.653 & 0.137 & \textbf{0.530} & 0.160 \\
GPT-4o      & \textbf{0.772} & 0.202 & 0.510 & 0.229 \\
o3          & 0.769 & \textbf{0.229} & 0.650 & \textbf{0.263} \\
\bottomrule
\end{tabular}
\caption{Vanilla prompting baseline}
\label{table:vanilla_comparison}
\end{table}

Notably, we observe a consistent progression in capability from GPT-4o-mini to GPT-4o and o3, highlighting the advancement in LLM-based reasoning even without task-specific tuning.

\subsection{Policy-Guided Test}

We now examine how predictive performance improves when the model is guided by the natural language policy we generated through our ICL procedure. For this evaluation, we used the most lightweight and cost-effective model available: GPT-4o-mini. The inference prompt used is as follows:

\begin{quote}
\ttfamily
You are an expert in venture capital, specializing in evaluating startup founders.\\
Your task is to distinguish successful founders from unsuccessful ones.\\
Here is a policy to assist you: \{policy\}

\vspace{0.5em}
Given the founder's profile:\\
Founder's LinkedIn Profile: \{row['clean\_linkedin\_profile']\}\\
Crunchbase information: \{row['clean\_cb\_profile']\}

\vspace{0.5em}
Based on this information, determine if the founder will succeed. Answer using only one word: True or False
\end{quote}

We tested two policy versions:
\begin{itemize}[itemsep=0.2em, topsep=0em]
    \item The \textit{initial policy}, derived from a small seed set and a single prompt
    \item The \textit{best-performing policy}, refined through multiple iterations of our ICL loop
\end{itemize}

The structure of the prompt remained the same, except the \texttt{policy} field was filled with either the initial or the best policy. The following are the evaluation results for the two policies:

\begin{table}[h!]
\centering
\begin{tabular}{l c c c c}
\toprule
Model & Accuracy & Precision & Recall & $F_{0.5}$ \\
\midrule
Initial Policy & 0.879 & 0.229 & 0.140 & 0.203 \\
Best Policy    & \textbf{0.917} & \textbf{0.645} & \textbf{0.200} & \textbf{0.446} \\
\bottomrule
\end{tabular}
\caption{Performance of initial and best-refined policy}
\label{table:policy_comparison}
\end{table}

These results demonstrate that an optimized policy substantially improves model performance, particularly in terms of precision and the overall $F_{0.5}$ score. In contrast, the initial policy provides only marginal gains over the vanilla model.

The improvement from the initial to the best policy highlights the effectiveness of our ICL loop. Despite using the lightweight model GPT-4o-mini, our method achieved results that clearly surpassed both the vanilla baselines and the early-stage policy. This suggests that the iterative refinement and scoring strategy we employ allows even smaller models to simulate domain expertise and apply learned reasoning patterns effectively.

\subsection{Test Stability and Results}
\label{sec:best_policy_eval}

To assess robustness, we tested our best-performing policy on eight distinct 100-success / 1000-failure test subsets. The results are shown below:

\begin{table}[h!]
\centering
\begin{tabular}{c c c c c}
\toprule
Test Set & Accuracy & Precision & Recall & $F_{0.5}$ \\
\midrule
1 & 0.917 & 0.645 & 0.20 & 0.446 \\
2 & 0.915 & 0.652 & 0.15 & 0.391 \\
3 & 0.913 & 0.583 & 0.14 & 0.357 \\
4 & 0.907 & 0.450 & 0.09 & 0.250 \\
5 & 0.912 & 0.552 & 0.16 & 0.370 \\
6 & 0.882 & 0.250 & 0.15 & 0.221 \\
7 & 0.870 & 0.205 & 0.15 & 0.191 \\
8 & 0.902 & 0.400 & 0.16 & 0.308 \\
\midrule
Mean & 0.902 & 0.467 & 0.15 & 0.317 \\
\bottomrule
\end{tabular}
\caption{Robustness of best-performing policy across eight test subsets}
\label{table:test_stability}
\end{table}

While overall performance is strong, we note that Test Sets 6 and 7 exhibit noticeably lower precision and $F_{0.5}$ scores. Further investigation using GPT-4o in vanilla (non-policy) mode confirmed that model performance on these subsets is inherently weaker, suggesting variability in data difficulty or distribution shift. Despite this, the mean precision across all eight test sets is approximately 0.467, indicating strong average performance of 5$\times$ and robustness to moderate dataset variation.

\subsection{Evaluation on Real-World Distribution}
\label{sec:section4.4}

VC firms typically review 30 to 50 startups per week, averaging around 2,000 per year. To simulate more realistic investment scenarios, we evaluated our best policy on four test sets that closely match real-world outlier\footnote{An outlier is often referred to as a \textit{unicorn} in the real world.} rates. Each test set contains 40 successful founders and 2,000 failed ones, approximating a 1.96\% success rate, which is the real-world random chance of success. 

Table~\ref{table:realistic_unicorn_test} shows that our model performs significantly better in relative terms on the more imbalanced dataset. Achieving an 8.8$\times$ lift over random precision suggests that the model is capable of identifying rare outliers under realistic startup success distributions. This demonstrates both robustness and practical utility in real-world venture scenarios.

\vspace{-0.5em}
\begin{table}[h]
\centering
\begin{tabular}{c c c c c}
\toprule
Test Set & Accuracy & Precision & Recall & $F_{0.5}$ \\
\midrule
1 & 0.975 & 0.308 & 0.200 & 0.278 \\
2 & 0.973 & 0.233 & 0.175 & 0.219 \\
3 & 0.953 & 0.077 & 0.125 & 0.083 \\
4 & 0.943 & 0.068 & 0.150 & 0.077 \\
\midrule
Mean & 0.961 & 0.172 & 0.163 & 0.164 \\
\bottomrule
\end{tabular}
\caption{Evaluation results on unicorn-base-rate test sets (40 success / 2000 failure)}
\label{table:realistic_unicorn_test}
\end{table}
\vspace{-0.5em}

\subsection{Finalizing the Policy with o3}

To explore the upper-bound potential of our approach, we employed o3 for the policy generation stage, leveraging its superior capabilities in logical reasoning and technical writing. For all other tasks, including case evaluation and scoring, we continued using GPT-4o-mini to maintain cost efficiency. This hybrid setup proved highly effective, yielding policies with significantly improved performance. The table below presents performance metrics on the same four test sets described in Section~\ref{sec:section4.4}.

\vspace{-0.5em}
\begin{table}[h]
\centering
\begin{tabular}{c c c c c}
\toprule
Test Set & Accuracy & Precision & Recall & $F_{0.5}$ \\
\midrule
1 & 0.981 & 0.600 & 0.150 & 0.375 \\
2 & 0.980 & 0.500 & 0.175 & 0.365 \\
3 & 0.976 & 0.250 & 0.100 & 0.192 \\
4 & 0.975 & 0.269 & 0.175 & 0.243 \\
\midrule
Mean & 0.978 & 0.405 & 0.150 & 0.294 \\
\bottomrule
\end{tabular}
\caption{Evaluation results on unicorn-base-rate test sets (40 success / 2000 failure)}

\end{table}

Notably, our final policy achieved an average precision of around 40\% on realistic test sets, representing more than 20$\times$ lift over random precision. These results underscore the potential of incorporating more advanced LLMs for policy induction. A targeted use of stronger models at critical stages can unlock significant downstream performance gains in founder evaluation.

\section{Training Details}

\subsection{Scoring System}\label{sec:scoring}
As described in Section~\ref{sec:icl-loop}, we adopted a simple but effective scoring system using the same data for both training and evaluation. Specifically, we evaluated candidate policies based on their precision on the training set. This approach is suitable because policies are general, language-based abstractions designed to capture broader reasoning patterns rather than memorize specific examples. Empirically, we found no significant difference in performance trends between using the training set and using a separate validation set. Thus, to reduce data, time, and cost requirements, we standardized our process to evaluate policy quality using the training set itself.

\subsection{Training Set Ratio}
We experimented with various success-to-failure ratios in the training data, including 1:5, 1:2, and 1:1. Our findings suggest that the absolute number of successful founders is more important than the overall ratio. Too few success examples reduce learning efficiency, leading to longer convergence times. A balanced 1:1 or lightly skewed ratio yielded the best trade-off between performance and training speed.

\subsection{Training Time}
A full training round using 120 data points (e.g., 60 success and 60 failure cases) with GPT-4o-mini typically takes 3--4 hours. Because policy generation and scoring involve multiple asynchronous LLM calls, training is I/O-bound and benefits from parallelization. Moreover, optimizing for high policy quality often requires many additional rounds of tuning and manual inspection, especially in the later stages of refinement.

\section{Conclusions}

Our methodology demonstrates that effective decision policies can be learned with as few as 120 training examples and no gradient-based updates, making it highly accessible. Through a looped ICL mechanism and lightweight scoring based on training precision, we achieved substantial gains in predictive power. The best policy achieved a mean precision of 0.405 across diverse test subsets, an improvement of more than 20$\times$ over the real-world precision baseline. This performance surpasses the estimated outlier-picking precision of top-tier VCs by 7.1$\times$.

The transparency of our approach also enables human experts to inspect, intervene, and contribute directly to policy refinement. This makes the system not only accurate but also trustworthy and adaptable. Because all decision logic remains in natural language, our framework can be ported to other domains that rely on structured reasoning, such as grant evaluation, academic hiring, or legal case triage.

Despite these promising results, several limitations remain. First, the training pipeline is inherently nondeterministic and sensitive to prompt phrasing and ordering, which can result in inconsistent outcomes across runs. Second, although we attempted to mitigate data contamination by excluding entity names, the LLMs may still retain latent exposure to parts of the dataset, potentially inflating the model's performance. Lastly, as seen in Test Sets 6 and 7, performance can vary substantially across different data segments, indicating that certain founder profiles remain challenging for LLMs to evaluate reliably.

In future work, we plan to explore multi-agent consensus generation, automated policy clustering, and integration with retrieval-augmented generation for more data-aware decision making. Our findings suggest that LLMs, when equipped with simple but powerful prompting strategies, can become practical, interpretable decision-making tools in real-world high-stakes domains such as venture capital.

\section*{Impact Statement}

This work contributes to the field of interpretable AI by providing a lightweight and explainable framework for high-stakes decision-making using LLMs. In particular, our method supports early-stage startup evaluation with a transparent and editable reasoning process. All datasets used in this study were collected and processed in accordance with relevant ethical standards, and no personally identifiable information was exposed to the model. We believe that this approach promotes responsible and auditable machine learning practices in financial domains.


\bibliography{main}
\bibliographystyle{icml2024}


\clearpage
\appendix
\section{Appendix: Best Policy and Experimental Details}

\subsection{Training Configuration}

The best-performing policy was generated using the following configuration:

\begin{table}[h]
\centering
\begin{tabular}{ll}
\toprule
\textbf{Training Round} & 4 \\
\textbf{Training Size} & 240 (120 success / 120 failure) \\
\textbf{Training Ratio} & 1:1 \\
\textbf{Training Range} & 0–120 (success and failure) \\
\textbf{Policy Length Limit} & 20 lines \\
\bottomrule
\end{tabular}
\caption{Training configuration for best policy}
\label{table:training_config}
\end{table}

\vspace{1em}

\subsection{Best Policy}

Below is the final version of the best-performing policy generated after four rounds of in-context refinement:

\textit{Updated Policies for Distinguishing Successful Founders:}
\begin{enumerate}
\item \textit{Industry Fit \& Scalability}: Prioritize founders building scalable tech, AI, or deep-tech products over service-heavy models.
\item \textit{Sector-Specific Innovation \& Patent Verification}: Require defensible IP with issued or published patents validated through public databases.
\item \textit{Quantifiable Outcomes, Exits \& (for Bio/Med) Regulatory Milestones}: Demand audited revenue, exits, or documented IND/clinical-phase progress—not just pre-clinical claims.
\item \textit{Funding \& Investor Validation}: Look for credible, recent third-party capital or follow-on rounds; stale or absent fundraising signals stagnation.
\item \textit{Press \& Recognition Depth}: Favor independent, reputable coverage within the last 24 months and cross-checked with filings; outdated or missing press is a red flag.
\item \textit{Product vs. Service Assessment}: Score higher for automated, high-margin SaaS, platform, or therapeutics with clear IP; pure services rank lower.
\item \textit{Market Traction Specificity}: Require cohort-level data on growth, retention, margins; name-dropping clients or “pilot” studies alone don’t qualify.
\item \textit{Location Advantage with Proof}: Presence in a tech/biotech hub must align with active local partnerships, accelerators, or ecosystem leadership roles.
\item \textit{Crisis Management \& Pivot History}: Validate data-backed pivots that preserved or grew value during downturns.
\item \textit{Sustainable 3--5-Year Roadmap}: Roadmap must tie to market trends, capital needs, and measurable milestones.
\item \textit{Skill Alignment \& Visibility}: Match proven technical, operational, or sales expertise to venture stage; generic “entrepreneur” labels penalize.
\item \textit{Consistent Role Tenure \& Title Concentration}: Favor $\geq$4-year focus in one core venture; multiple simultaneous C-suite/advisory titles or role inflation is a downgrade.
\item \textit{Network Quality \& Engagement}: Measure depth and actual engagement of investor and domain-expert ties over raw connection counts.
\item \textit{Third-Party Validation \& References}: Require testimonials, case studies, regulatory filings, or audits corroborating performance and scientific claims.
\item \textit{Investment Ecosystem Participation}: Credit active, recent angel or fund roles that demonstrate curated deal flow and learning loops.
\item \textit{Differentiated Value Proposition}: Demand a clear, data-supported statement of competitive advantage and defensibility.
\item \textit{Tech Currency \& Relevance}: Ensure the founder’s expertise, tech stack, and go-to-market playbook are current; legacy success alone is insufficient.
\item \textit{Data Consistency Across Platforms}: Cross-verify LinkedIn, Crunchbase, press, and regulatory filings; inconsistencies or absent data trigger deeper diligence or rejection.
\end{enumerate}

\vspace{1em}

\subsection{Vanilla GPT-4o Performance on Difficult Sets}

To better understand the performance gap, we ran GPT-4o in vanilla (no policy) mode on Test Sets 1 (for comparison), 6 and 7:

\begin{itemize}
\item \textbf{Test Set 1 (Vanilla GPT-4o)}: Precision = 0.202, $F_{0.5}$ = 0.229
\item \textbf{Test Set 6 (Vanilla GPT-4o)}: Precision = 0.153, $F_{0.5}$ = 0.178
\item \textbf{Test Set 7 (Vanilla GPT-4o)}: Precision = 0.134, $F_{0.5}$ = 0.157
\end{itemize}

These results confirm that the lower scores for Sets 6 and 7 are not unique to our policy-based model and may reflect structural difficulty in those segments.




\end{document}